\documentclass[letterpaper,10pt,conference]{ieeeconf}

\IEEEoverridecommandlockouts
\let\labelindent\relax

\usepackage{url}

\usepackage{color}
\usepackage{amsmath}
\usepackage{todonotes}
\usepackage{xspace}
\usepackage{enumitem}
\usepackage{algorithm}
\usepackage{algpseudocode}
\usepackage{amsfonts}
\usepackage{cite}
\usepackage{subcaption}
\usepackage[hidelinks]{hyperref}
\usepackage{nameref}

\usepackage{booktabs}
\usepackage{multirow}
\usepackage{dcolumn}
\newcolumntype{d}[1]{D{.}{.}{#1}}
\usepackage[normalem]{ulem}
\usepackage{comment}
\usepackage{amsmath}

\DeclareMathOperator*{\argmin}{arg\,min}

\usepackage{xcolor}

\definecolor{darkgreen}{RGB}{100, 100, 0}

\newtheorem{proof}{Proof}

\usepackage{pifont}
\newcommand{\cmark}{\ding{51}}%
\newcommand{\xmark}{\ding{55}}%

\graphicspath{{figs/}}

\begin{document}

\title{\LARGE \bf Pred-NBV: Prediction-guided Next-Best-View Planning\\ for 3D Object Reconstruction}

\author{ Harnaik Dhami* \and Vishnu D. Sharma* \and Pratap Tokekar
\thanks{*Equal contribution. Names are listed alphabetically.}
\thanks{Authors are with the Department of Computer Science, University of Maryland, U.S.A. \texttt{\small \{dhami, vishnuds, tokekar\}@umd.edu}.}\thanks{This work is supported by the National Science Foundation under grant number 1943368 and ONR under grant number N00014-18-1-2829.}}

\maketitle
\IEEEpeerreviewmaketitle

\begin{abstract}
    Prediction-based active perception has shown the potential to improve the navigation efficiency and safety of the robot by anticipating the uncertainty in the unknown environment. The existing works for 3D shape prediction make an implicit assumption about the partial observations and therefore cannot be used for real-world planning and do not consider the control effort for next-best-view planning. We present \textit{Pred-NBV}, a realistic object shape reconstruction method consisting of PoinTr-C, an enhanced 3D prediction model trained on the ShapeNet dataset, and an information and control effort-based next-best-view method to address these issues. \textit{Pred-NBV} shows an improvement of \textbf{25.46\%} in object coverage over the traditional methods in the AirSim simulator, and performs better shape completion than PoinTr, the state-of-the-art shape completion model, even on real data obtained from a Velodyne 3D LiDAR mounted on DJI M600 Pro.
\end{abstract}

\section{Introduction}

The goal of this paper is to improve the efficiency of mapping and reconstructing an object of interest with a mobile robot. This is a long studied and fundamental problem in the field of robotics~\cite{bajcsy2018revisiting}. In particular, the commonly used approach is Next-Best-View (NBV) planning. In NBV planning, the robot seeks to find the best location to go to next and obtain sensory information that will aid in reconstructing the object of interest. A number of approaches for NBV planning have been proposed over the years~\cite{delmerico2018comparison}. In this paper, we show how to leverage the recent improvements in perception due to deep learning to improve the efficiency of 3D object reconstruction with NBV planning. In particular, we present a 3D shape prediction technique that can predict a full 3D model based on the partial views of the object seen so far by the robot to find the NBV. Notably, our method works ``in the wild'' by eschewing some common assumptions made in 3D shape prediction, namely, assuming that the partial views are still centered at the full object center.

There are several applications where robots are being used for visual data collection. Some examples include  inspection for visual defect identification of civil infrastructure such as bridges~\cite{shanthakumar2018view,dhamiGATSBI}, ship hulls~\cite{kim_2009_IROS} and aeroplanes~\cite{ropek_2021}, digital mapping for real estate~\cite{46965,ramakrishnan2021hm3d}, and precision agriculture~\cite{dhami2020crop}. The key reasons why robots are used in such applications is that they can reach regions that are not easily accessible for humans and we can precisely control where the images are taken from. However, existing practices for the most part require humans to specify a nominal trajectory for the robots that will visually cover the object of interest. Our goal in this paper is to automate this process.  

\begin{figure}[ht!]
    \centering
    \includegraphics[width=\linewidth]{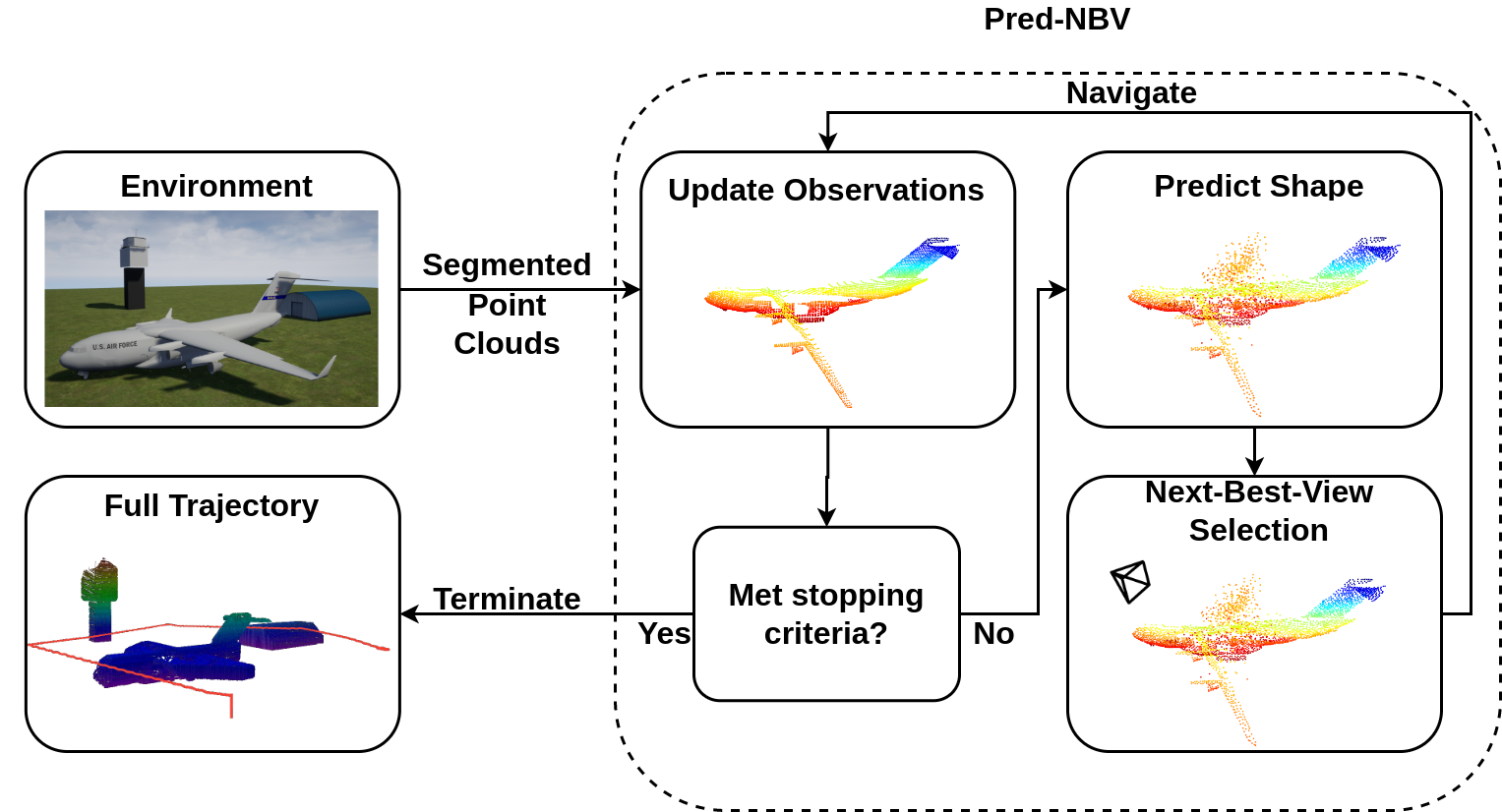}
    \caption{Overview of the proposed approach}
    \label{fig:overview}
\end{figure}

The NBV planning method is the commonly used approach to autonomously decide where to obtain the next measurement from. NBV planning typically uses geometric cues such as symmetry~\cite{debevec1996modeling} or prior information~\cite{breyer2022closed} for deciding the next best location. In this work, we do not rely on such assumptions but instead leverage the predictive power of deep neural networks for 3D shape reconstruction.

Recent works have explored predictions as a way of improving these systems by anticipating the unknowns with prediction and guiding robots' motion accordingly. This approach has been studied for robot navigation, exploration, and manipulation~\cite{ramakrishnan2020occupancy,georgakis2022uncertainty,Wei2021,yan2019data} with the help of neural network-based methods that learn from datasets. While 2D map presentation works have shown the benefits of robotic tasks in simulation as well as the real-world, similar methods for 3D prediction have been limited to simplistic simulations. The latter approaches generally rely on synthetic datasets due to the lack of realistic counterparts for learning.

Strong reliance on data results in the neural network learning implicit biases. Some models can make predictions only for specific objects~\cite{3drecgan}. Others require implicit knowledge of the object's center, despite being partially visible~\cite{yu2021pointr}. These situations are invalid in real-world, mapless scenarios and may result in inaccurate shape estimation. Many shape prediction works assume the effortless motion of the robot~\cite{wu20153d}, whereas an optimal path for a robot should include the effort required to reach a position and the potential information gain due to time and power constraints. Monolithic neural networks replacing both the perception and planning components present an alternative, but they tend to be specific to the datasets used for training and may require extensive finetuning for real-world deployment. Lack of transparency in such networks poses another challenge that could be critical for the safe operation of a robot when working alongside humans.

To make 3D predictive planning more realistic, efficient, and safe, we propose a method consisting of a 3D point cloud completion model, that relaxes the assumption about implicit knowledge of the object's center using a curriculum learning framework~\cite{bengio2009curriculum}, and an NBV framework, that maximizes the information gain from image rendering and minimizes the distance traveled by the robot. Furthermore, our approach is modular making it interpretable and easy to upgrade.

We make the following contributions to this work:
\begin{itemize}
    \item We use curriculum learning to build an improved 3D point cloud completion model, which does not require the partial point cloud to be centered at the full point cloud's center and is more robust to perturbations than earlier models. We show that this model, termed \textit{PoinTr-C}, outperforms the base model, PoinTr~\cite{yu2021pointr}, by at least $\textbf{23.06}\%$ and show qualitative comparison on ShapeNet~\cite{shapenet2015} dataset, and real point cloud obtained with a Velodyne 3D LiDAR mounted on DJI M600 Pro.
    \item We propose a next-best-view planning approach that performs object reconstruction without any prior information about the geometry, using predictions to optimize information gain and control effort over a range of objects in a model-agnostic fashion.
    \item We show that our method covers on average \textbf{25.46\%} more points on all models evaluated for object reconstruction in AirSim~\cite{airsim2017fsr} simulations compared to the non-predictive baseline approach, \textit{Basic-Next-Best-View}~\cite{aleotti2014global} and performs even better for complex structures such as airplanes.
\end{itemize}

We share the qualitative results, project code, and visualization from our method on our project website\footnote{Project webpage: \url{http://raaslab.org/projects/PredNBV/}}.

\section{Related Work}\label{sec:rel_work}

Active reconstruction in an unknown environment can be accomplished through next-best-view (NBV) planning, which has been studied by the robotics community for a long time~\cite{scott2003view}. In this approach, the robot builds a partial model of the environment based on observations and then moves to a new location to maximize the cumulative information gained. The NBV approaches can be broadly classified into information-theoretic and geometric methods. The former builds a probabilistic occupancy map from the observations and uses the information-theoretic measure~\cite{delmerico2018comparison} to select the NBV. The latter assumes the partial information to be exact and determines the NBV based on geometric measures~\cite{tarabanis1995survey}.

The existing works on NBV with robots focus heavily on  information-theoretic approaches for exploration in 2D and 3D environments~\cite{kuipers1991robot,vasquez2014volumetric}. Subsequent development for NBV with frontier and tree-based approaches was also designed for exploration by moving the robot towards unknown regions~\cite{yamauchi1997frontier, gonzalez2002navigation, adler2014autonomous, bircher2018receding}. Prior works on NBV for object reconstruction also rely heavily on information-theoretic approaches to reduce uncertainty in pre-defined closed spaces~\cite{morooka1998next, vasquez2009view}. Geometric approaches require knowing the model of the object in some form and thus have not been explored to a similar extent.
Such existing works try to infer the object geometry from a database or as an unknown closed shape~\cite{banta2000next, kriegel2013combining}, and thus may be limited in application.

In recent years, prediction-based approaches have emerged as another solution. One body of these approaches works in conjunction with other exploration techniques to improve exploration efficiency by learning to predict structures in the environment from a partial observation. This is accomplished by learning the common structures in the environment (buildings and furniture, for example) from large datasets. This approach has recently gained traction and has been shown to work well for mobile robots navigation with 2D occupancy map representations~\cite{ramakrishnan2020occupancy, Wei2021, sharma2023proxmap}, exploration~\cite{georgakis2022uncertainty}, high-speed maneuvers~\cite{Katyal2021}, and elevation mapping~\cite{yang2022real}.

Similar works on 3D representations have focused mainly on prediction modules. Works along this line have proposed generating 3D models from novel views using single RGB image input~\cite{Hani2020}, depth images~\cite{Yang_2019}, normalized digital surface models (nDSM)~\cite{Alidoost2019}, point clouds~\cite{yu2021pointr, xie2020grnet, yuan2018pcn}, etc. The focus of these works is solely on inferring shapes based on huge datasets of 3D point clouds~\cite{shapenet2015}. They do not discuss the downstream task of planning. A key gap in these works is that they assume a canonical representation of the object, such as the center of the whole object, to be provided either explicitly or implicitly. Relaxing this assumption does not work well in the real-world where the center of the object may not be estimated accurately, discouraging the adoption of 3D prediction models for prediction-driven planning.

Another school of work using 3D predictions combines the perception and planning modules as a neural network. These works, aimed at predicting the NBV to guide the robot from partial observations, were developed for simple objects~\cite{POP2022160}, 3D house models~\cite{Peralta2020}, and a variety of objects~\cite{zeng2020pc} ranging from remotes to rockets. Peralta et al.~\cite{Peralta2020} propose a reinforcement-learning framework, which can be difficult to implement due to sampling complexity issues. The supervised-learning approach proposed by Zeng et al.~\cite{zeng2020pc} predicts the NBV using a partial point cloud, but the candidate locations must lie on a sphere around the object, restricting the robot's planning space. Moreover, monolithic neural networks suffer from a lack of transparency and real-world deployment may require extensive fine-tuning of the hyperparameters. Prediction-based modular approaches solve these problems as the intermediate outputs are available for interpretation and the prediction model can be plugged in with the preferred planning method for a real environment.

A significant contribution of our work is to relax the implicit assumption used in many works that the center and the canonical orientation of the object under consideration are known beforehand, even if the 3D shape completion framework uses partial information as the input. A realistic inspection system may not know this information and thus the existing works may not be practically deployable.

\section{Problem Formulation}\label{sec:prob_form}

We are given a robot with a 3D sensor onboard that explores a closed object with volume $\mathcal{V} \in \mathbb{R}^3$. The set of points on the surface of the object is denoted by $\mathcal{S} \in \mathbb{R}^3$. The robot can move in free space around the object and observe its surface. The surface of the object $s_i \subset \mathcal{S}$ perceived by the 3D sensor from the pose $\phi_i \subset \Phi$ 

is represented as a voxel-filtered point cloud. We define the relationship  between the set of points observed from a view-point $\phi_i$ with a function $f$, i.e., $s_i = f(\phi_i)$. The robot can traverse  a trajectory $\xi$ that consists of view-points $\{\phi_1, \phi_2, \ldots, \phi_m \}$. The surface observed over a trajectory is the union of surface points observed from the consisting viewpoints, i.e. $s_{\xi} = \bigcup_{\phi \in \xi} f(\phi)$. The distance traversed by the robot between two view-points $\phi_i$ and $\phi_j$ is denoted by $d(\phi_i, \phi_j)$.

Our objective is to find a trajectory $\xi_i$ from the set of all possible trajectories $\Xi$, such that it observes the whole voxel-filtered surface

of the object while minimizing the distance traversed.  
\begin{align}
    \xi^* = \argmin_{\xi \in \Xi} \sum_{i=1}^{| {\xi}| - 1} d(\phi_i, \phi_{i+1}),~ 
    \textit{such that} \bigcup_{\phi_i \in \xi} f(\phi_i) = \mathcal{S}.
\end{align}

In unseen environments, $\mathcal{S}$ is not known apriori, hence the optimal trajectory can not be determined. We assume that the robot starts with a view of the object. If not, we can always first explore the environment until the object of interest becomes visible. 

\section{Proposed Approach}\label{sec:approach}
We propose \textit{Pred-NBV}, a prediction-guided NBV method for 3D object reconstruction highlighted in Fig.~\ref{fig:overview}. Our method consists of two key modules: (1) \textit{PoinTr-C}, a robust 3D prediction model that completes the point cloud using only partial observations, and (2) an NBV framework that uses prediction-based information gain to reduce the control effort for active object reconstruction. We provide the details in the following subsection.

\vspace{-1mm}
\subsection{\textit{PoinTr-C}: 3D Shape Completion Network}
\vspace{-0.5mm}
Given the current set of observations $v_o \in \mathcal{V}$, we predict the complete volume using a learning-based predictor $g$, i.e., $\hat{\mathcal{V}} = g(v_o)$.

To obtain $\hat{\mathcal{V}}$, we use PoinTr~\cite{yu2021pointr}, a transformer-based architecture that uses 3D point clouds as the input and output. 

PoinTr uses multiple types of machine learning methods to perform shape completion. It first identifies the geometric relationship in  low resolution between points in the cloud by clustering. Then it generates features around the cluster centers, which are then fed to a transformer~\cite{vaswani2017attention} to capture the long-range relationships and predict the centers for the missing point cloud. Finally, a coarse-to-fine transformation over the predicted centers using a neural network outputs the missing point cloud. 

This model was trained on the ShapeNet~\cite{shapenet2015} dataset and outperforms the previous methods on a range of objects. However, PoinTr was trained with implicit knowledge of the center of the object. Moving the partially observed point cloud to its center results in incorrect prediction from PoinTr.

To improve predictions, we fine-tune PoinTr using a curriculum framework, which dictates training the network over easy to hard tasks by increasing the difficulty in steps during learning~\cite{bengio2009curriculum}. Specifically, we fine-tune PoinTr over increasing perturbations in rotation and translation to the canonical representation of the object to relax the assumption about implicit knowledge of the object's center. We use successive rotation-translation pairs of $(25^\circ, 0.0)$, $(25^\circ, 0.1)$, $(45^\circ, 0.1)$, $(45^\circ, 0.25)$, $(45^\circ, 0.5)$, $(90^\circ, 0.5)$, $(180^\circ, 0.5)$, and $(360^\circ, 0.5)$  for curriculum training.
We assume that the object point cloud is segmented well, which can be achieved using distance-based filters or segmentation networks.

\vspace{-1mm}
\subsection{Next-Best View Planner}
\vspace{-0.5mm}
Given the predicted point cloud $\mathcal{\hat{V}}$ for the robot after traversing the trajectory $\xi_{t}$, we generate a set of candidate poses $\mathcal{C} = \{ \phi_1, \phi_2, ..., \phi_m \}$ around the object observed so far. Given $v_o$, the observations so far, we define the objective to select the shortest path that results in observing at least $\tau \%$ of the maximum possible information gain over all the candidate poses. Considering $\mathcal{\hat{V}}$ as an exact model, we use a geometric measure to quantify the information gained from the candidate poses. Specifically, we define a projection function $I(\xi)$, over the trajectory $\xi$, which first identifies the predicted points distinct from the observed point cloud over the trajectory, then apply a hidden point removal operator on them~\cite{katz2007direct}, without reconstructing a surface or estimating a normal, and lastly, find the number of points that will be observed if we render an image on the robot's camera. Thus, we find the NBV from the candidate set $\mathcal{C}$ as follows:
\begin{align*}
    \phi_{t+1} = \argmin_{\phi \in \mathcal{C}} d(\phi, \phi_{t}),
    ~\textit{such that}~ \frac{I(\xi_t \cup \phi)}{\max_{\phi \in \mathcal{C}} I(\xi_t \cup \phi)} \geq \tau.
\end{align*}

We find the $d(\phi_i, \phi_j)$, using RRT-Connect~\cite{kuffner2000rrt}, which incrementally builds two rapidly-exploring random trees rooted at $\phi_i$ and $\phi_j$ through the observed space to provide a safe trajectory. After selecting the NBV, the robot follows this trajectory to reach the prescribed view-point. We repeat the prediction and planning process until the ratio of observations in the previous step and current step is $0.95$ or higher.

To generate the candidate set $\mathcal{C}$, we first find the distance $d_{max}$ of the point farthest from the center of the predicted point cloud $\mathcal{\hat{V}}$ and z-range. Then, we generate candidate poses on three concentric circles: one centered at $\mathcal{\hat{V}}$ with radius $1.5 \times d_{max}$ at steps of $30^\circ$, and one $0.25 \times \text{z-range}$ above and below with radius $1.2 \times d_{max}$ at steps of $30^\circ$. We use $\tau = 0.95$ for all our experiments.

\section{Evaluation}\label{sec:eval}
In this section we evaluate the \textit{Pred-NBV} pipeline. We start with a qualitative example followed by a comparison of the individual modules against respective baseline methods. The results show that \textit{Pred-NBV} is able to outperform the baselines significantly using large-scale models from the Shapenet~\cite{shapenet2015} dataset and with real-world 3D LiDAR data.

\subsection{Qualitative Example}
\begin{figure}[ht!]
    \centering
    \begin{subfigure}[b]{.7\columnwidth}%
        \includegraphics[width = \textwidth]{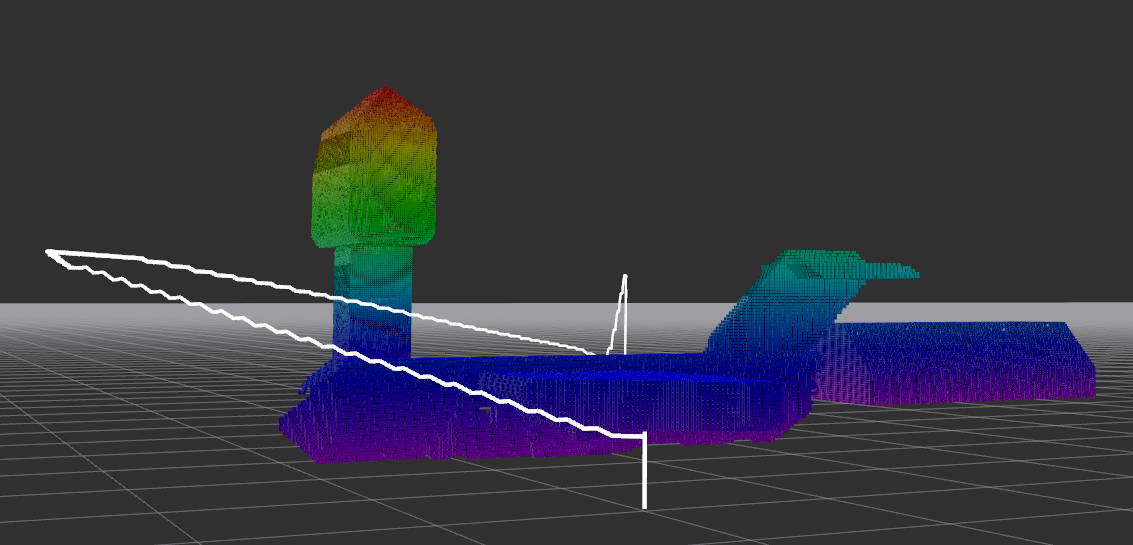}%
    \end{subfigure}%
    \hfill%
    \begin{subfigure}[b]{.5\columnwidth}%
        \includegraphics[width = \textwidth]{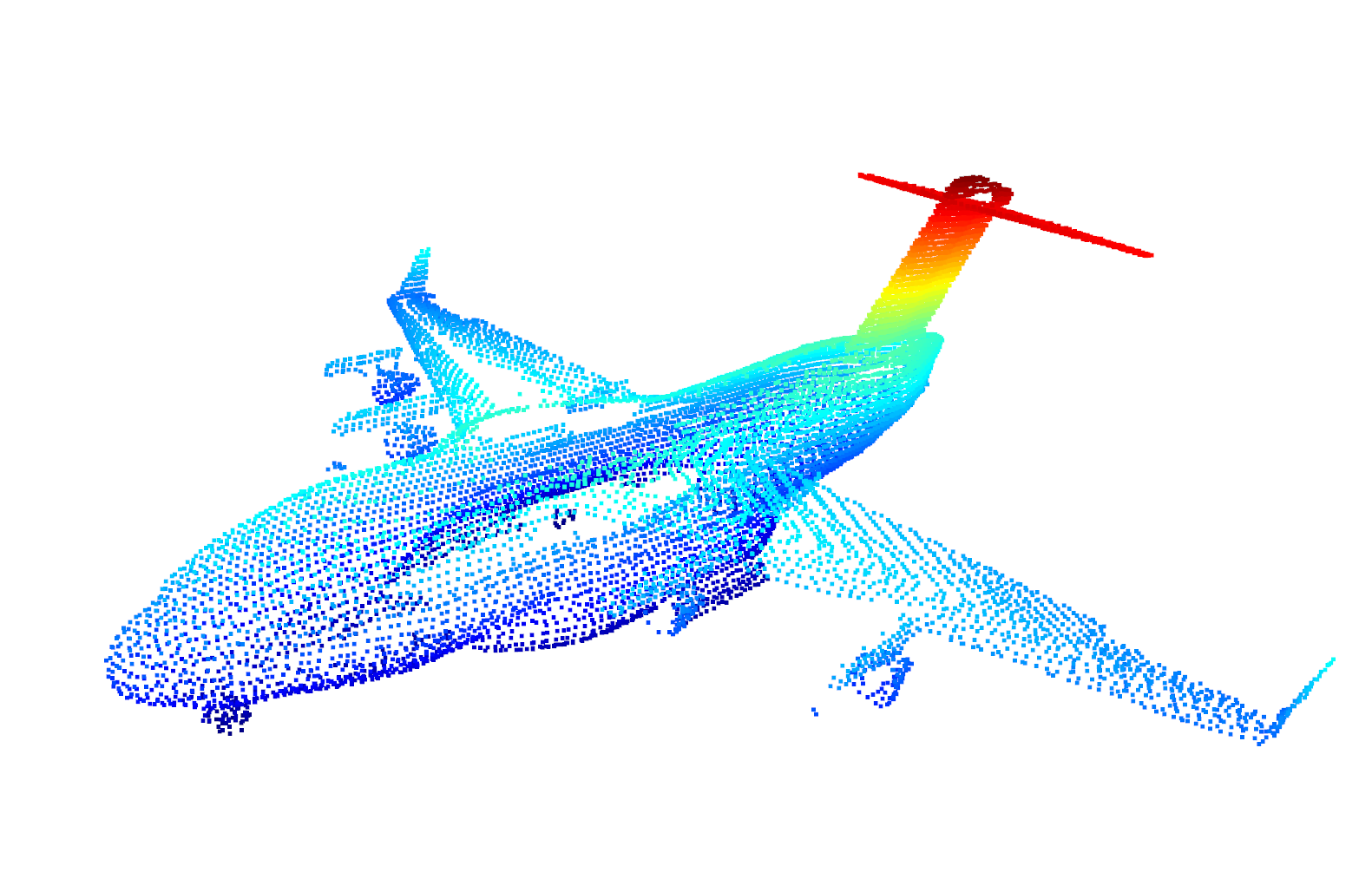}%
    \end{subfigure}%
    \caption{Flight path and total observations of C17 Airplane after running our NBV planner in AirSim simulation.}
    \label{fig:observations_airsim}
    \vspace{-5mm}
\end{figure}

\vspace{-0.5mm}
 Fig.~\ref{fig:observations_airsim} show the reconstructed point cloud of a C17 airplane and the path followed by a UAV in AirSim~\cite{airsim2017fsr}. We create candidate poses on three concentric rings at different heights around the center of the partially observed point cloud. The candidate poses change as more of the object is visible. 

 As shown in Fig.~\ref{fig:plane_res}, \textit{Pred-NBV} observes more points than the NBV planner without prediction in the same time budget. 

\vspace{-1mm}
\subsection{3D Shape Prediction}
\vspace{-0.5mm}
\subsubsection{Setup}

We train \textit{PoinTr-C} on a 32-core, 2.10Ghz Xeon Silver-4208 CPU and Nvidia GeForce RTX 2080Ti GPU with 11GB of memory. The network is fine-tuned over the ShapeNet\cite{shapenet2015} dataset, trained with perturbation as described in Section~\ref{sec:approach}. Similar to PoinTr~\cite{yu2021pointr}, we use Chamfer distance (CD) and Earth Mover's Distance (EMD), permutation-invariant metrics suggested by Fan et al.~\cite{fan2017point}, as the loss function for training \textit{PoinTr-C}. For evaluation we use two versions of Chamfer distance: CD-$l_1$ and CD-$l_2$, which use L1 and L2-norm, respectively, to calculate the distance between two sets of points, and F-score which quantifies the percentage of points reconstructed correctly.

\subsubsection{Results}
Table~\ref{tab:base_vs_best_mean} summarizes our findings regarding the effect of perturbations. \textit{PoinTr-C} outperforms the baseline in both scenarios. It only falters in CD-$l_2$ in the ideal condition, i.e., no augmentation. Furthermore, \textit{PoinTr-C} doesn't undergo large changes in the presence of augmentations, making it more robust than PoinTr. The relative improvement for \textit{PoinTr-C} at least $23.05\%$ (F-Score).

\begin{table}[h]
    \centering
    \vspace{-2.5mm}
    \caption{Comparison between the baseline model (PoinTr) and \textit{PoinTr-C} over test data with and without perturbation. Arrows show if a higher ($\uparrow$) or a lower ($\downarrow$) value is better.}
    \begin{tabular}{llrrr}
        \toprule
        Perturbation  & Approach  &  F-Score $\uparrow$   &  CD-$l_1$ $\downarrow$  & CD-$l_2$ $\downarrow$\\
        \midrule
        \multirow{2}*{\xmark} & PoinTr\cite{yu2021pointr} & 0.497 & 11.621 & \textbf{0.577}\\
        {} & \textit{PoinTr-C} & \textbf{0.550} & \textbf{10.024} & 0.651\\
        \midrule
        \multirow{2}*{\cmark} & PoinTr\cite{yu2021pointr} & 0.436 & 16.464 & 1.717\\
        {} & \textit{PoinTr-C} & \textbf{0.550} & \textbf{10.236} & \textbf{0.717}\\
        \bottomrule
        \end{tabular}
        \vspace{-2.5mm}
        \label{tab:base_vs_best_mean}
\end{table}

\begin{figure}
    \centering
    \vspace{0mm}
    \begin{subfigure}[b]{0.25\columnwidth}%
        \includegraphics[width=\linewidth]{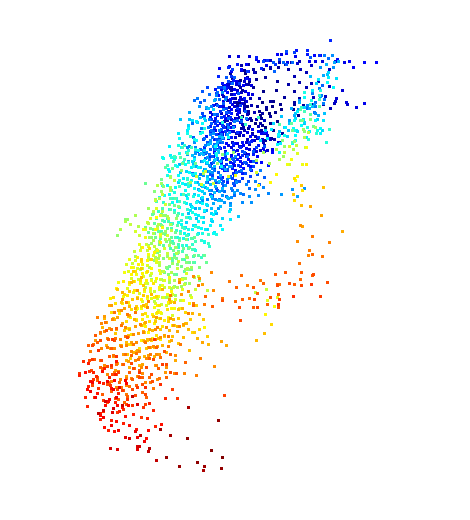}
        \subcaption{Input}
    \end{subfigure}%
    \hfill%
    \begin{subfigure}[b]{0.25\columnwidth}%
        \includegraphics[width=\linewidth]{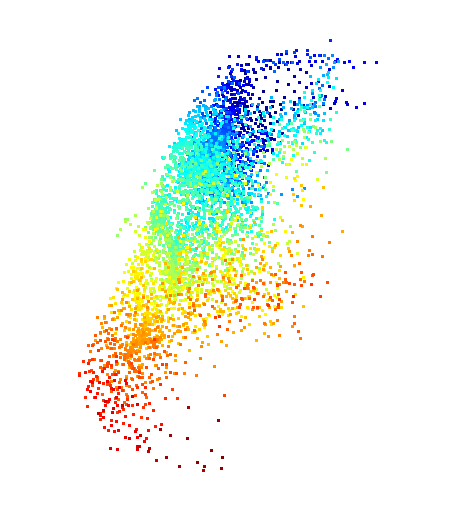}
        \subcaption{PoinTr~\cite{yu2021pointr}}
    \end{subfigure}%
    \hfill%
    \begin{subfigure}[b]{0.25\columnwidth}%
        \includegraphics[width=\linewidth]{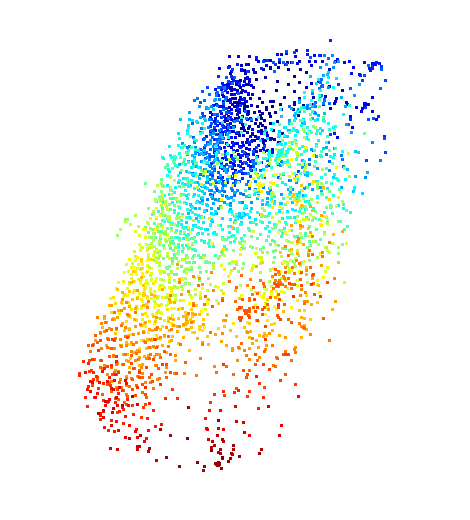}
        \subcaption{PoinTr-C}
    \end{subfigure}%
    \caption{Results over the real-world point cloud of a car obtained using LiDAR (Interactive figure available on \href{http://raaslab.org/projects/PredNBV/}{our webpage}).}
    \label{fig:qualitative_real_colored}
\end{figure}

We provide a qualitative comparison of the predictions from the two models for various objects under perturbations on our \href{http://raaslab.org/projects/prednbv/}{project webpage}. Fig.~\ref{fig:qualitative_real_colored} shows the results for a real point cloud of a car (visualized from the camera above the left headlight) obtained with a Velodyne LiDAR sensor. The results show the predictions from PoinTr are scattered around the center of the visible point cloud, whereas \textit{PoinTr-C} makes more realistic predictions. Our webpage provides interactive visualizations of these point 
clouds for further inspection.

\vspace{-1mm}
\subsection{Next-Best-View Planning}
\vspace{-0.5mm}
\subsubsection{Setup}

We use Robot Operating System (ROS) Melodic and AirSim~\cite{airsim2017fsr} on Ubuntu 18.04 to carry out the simulations. We equipped the virtual UAV with a RGBD camera. AirSim's built-in image segmentation is used to segment out the target object from the rest of the environment. We created a ROS package to publish a segmented depth image containing pixels only belonging to the bridge based on the RGB camera segmentation. This segmented depth image was then converted to a point cloud. We use the MoveIt~\cite{coleman2014reducing} software package based on the work done by Köse~\cite{tahsinko86:online} to implement the RRT connect algorithm. MoveIt uses RRT connect and the environmental 3D occupancy grid to find collision-free paths for point-to-point navigation.

\subsubsection{Qualitative Example}

We evaluate \textit{Pred-NBV} on 20 objects from 5 ShapeNet classes:  airplane, rocket, tower, train, and watercraft. We selected these classes as they represent larger shapes suitable for inspection. Fig.~\ref{fig:overview} shows the path followed by the UAV using \textit{Pred-NBV} for the C-17 airplane simulation. There are non-target obstacles in the environment, such as a hangar and air traffic control tower. \textit{Pred-NBV} finds a collision-free path that selects viewpoints targeting maximum coverage of the airplane.

\subsubsection{Comparison with Baseline}\label{sec:sim:baseline}
We compare the performance of \textit{Pred-NBV} with a baseline NBV method~\cite{aleotti2014global}. The baseline selects poses based on frontiers in the observed space using occupancy grids. We modified the baseline to improve it for our application and make it comparable to \textit{Pred-NBV}. The modifications include using our segmentation for the occupancy grid so that frontiers are weighted towards the target object. We also set the orientation of the selected poses towards the center of the target object similar to how \textit{Pred-NBV} works. The algorithms had the same stopping criteria as \textit{Pred-NBV}.

\begin{figure}
    \vspace{1.3mm}
    \centering
    \begin{subfigure}[b]{\columnwidth}%
        \hspace{0.8cm}
        \includegraphics[width = .75\textwidth]{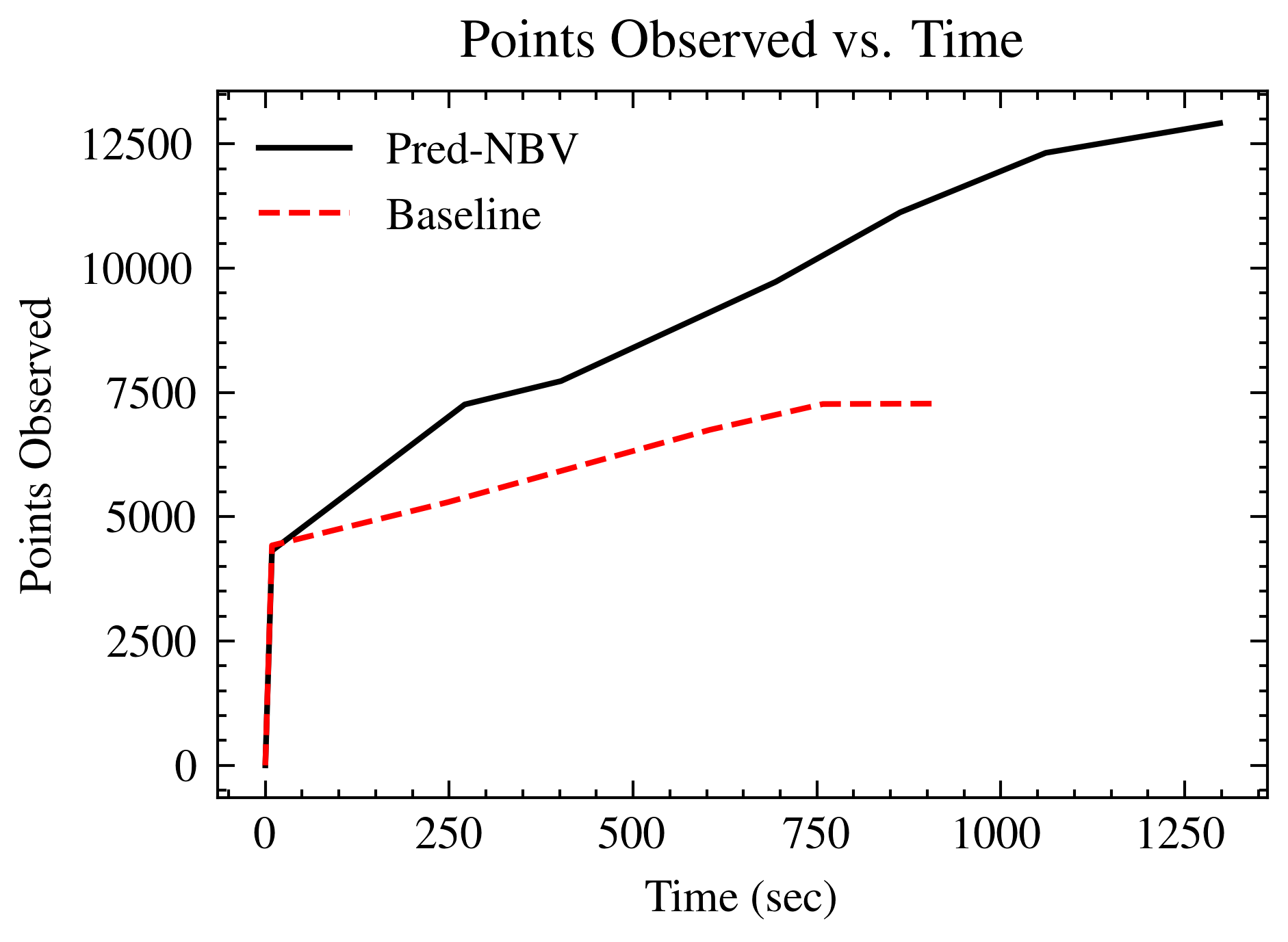}%
    \end{subfigure}%
    \hfill%
    \begin{subfigure}[b]{\columnwidth}%
        \hspace{1.0cm}
        \includegraphics[width = .725\textwidth]{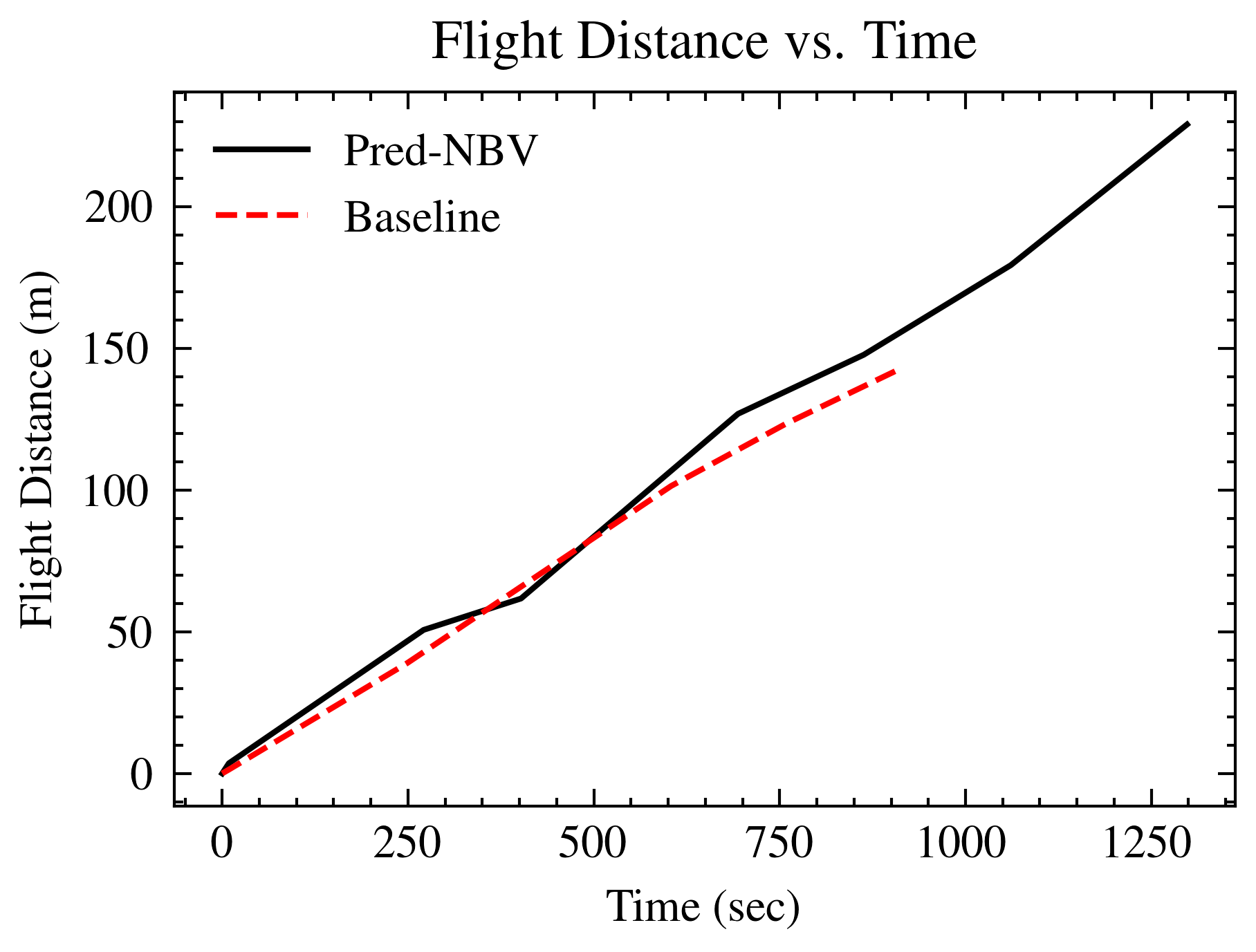}%
    \end{subfigure}%
    \caption{Comparison between \textcolor{black}{\textbf{Pred-NBV}} and the \textcolor{red}{\textbf{baseline NBV algorithm}}~\cite{aleotti2014global} for a C-17 airplane.}
    \label{fig:plane_res}
\end{figure}

We see in Table~\ref{tab:airsim_results} that our method observes on average 25.46\% more points than the baseline for object reconstruction across multiple models from various classes. In Fig.~\ref{fig:plane_res}, we show that \textit{Pred-NBV} observes more points per step than the baseline while not flying further per each step. 

\begin{table}[ht!]
    \centering
    \vspace{1.3mm}
    \caption{Points observed by \textit{Pred-NBV} and the baseline NBV method~\cite{aleotti2014global} for all models in AirSim.}
    \begin{tabular}{llrrr}
    
        \toprule
        \multirow{2}{*}{ Class} & \multirow{2}{*}{Model} & Points Seen & Points Seen & \multirow{2}{*}{Improvement} \\
        & & \textit{Pred-NBV} & Baseline & \\
        \midrule
        \multirow{5}{*}{Airplane}  
            & 747 & \textbf{11922} & 9657 & 20.99\% \\
            & A340 & \textbf{8603} & 5238 &  48.62\% \\
            & C-17 & \textbf{12916} & 7277 & 55.85\% \\
            & C-130 & \textbf{9900} & 7929 & 22.11\% \\
            & Fokker 100 & \textbf{10192} & 9100 & 11.32\%\\
        \midrule
        \multirow{5}{*}{Rocket} 
            & Atlas & \textbf{1822} & 1722 & 5.64\% \\
            & Maverick & \textbf{2873} & 2643 & 8.34\% \\
            & Saturn V & \textbf{1111} & 807 & 31.70\% \\
            & Sparrow & \textbf{1785} & 1639 & 8.53\% \\
            & V2 & \textbf{1264} & 1086 & 15.15\% \\
        \midrule
        \multirow{5}{*}{Tower} 
            & Big Ben & \textbf{4119} & 3340 & 20.89\% \\
            & Church & \textbf{2965} & 2588 & 13.58\% \\
            & Clock & \textbf{2660} & 1947 & 30.95\% \\
            & Pylon & \textbf{3181} & 2479 & 24.80\% \\
            & Silo & \textbf{5674} & 3459 & 48.51\% \\
        \midrule
        \multirow{2}{*}{Train}
            & Diesel & \textbf{3421} & 3161 & 7.90\% \\
            & Mountain & \textbf{4545} & 4222 & 7.37\% \\
        \midrule
        \multirow{3}{*}{Watercraft} 
            & Cruise & \textbf{4733} & 3522 & 29.34\% \\
            & Patrol & \textbf{3957} & 2306 & 52.72\% \\
            & Yacht & \textbf{9499} & 6016 & 44.90\% \\
        \bottomrule
    \end{tabular}
    \label{tab:airsim_results}
\end{table}

\vspace{-5mm}
\section{Conclusion}\label{sec:con}
We propose a realistic and efficient planning approach for robotic inspection using learning-based predictions. Our approach fills the gap between the existing works and the realistic setting by proposing a curriculum-learning-based point cloud prediction model, and a distance and information gain aware inspection planner for efficient operation. Our analysis shows that our approach is able to outperform the baseline approach in observing the object surface by 25.46\%. Furthermore, we show that our predictive model is able to provide satisfactory results for real-world point cloud data. We believe the modular design paves the path to further improvement by enhancement of the constituents. 

In this work we use noise free observations, but show that \textit{Pred-NBV} has the potential to work well on real, noisy inputs with pre-processing. In future work, we will explore making the prediction network robust to noisy inputs and with implicit filtering capabilities. We used a geometric measure for NBV in this work, and will extend it to information-theoretic measures using ensemble of predictions and uncertainty extraction techniques in future work.

{\small
\bibliographystyle{IEEEtran}
\bibliography{main}
}

\end{document}